%% file: paper.tex
\def\myauthor{
Robert J. Griffin\textsuperscript{1,2}, 
Sylvain Bertrand\textsuperscript{1},
Georg Wiedebach\textsuperscript{1},
Alexander Leonessa\textsuperscript{2},
Jerry Pratt\textsuperscript{1}
\thanks{\footnotesize{This work was funded through the NSF NRI Grant No. 1525972 and by the NASA Grant No. NNX12AP97G.}}
\thanks{\footnotesize{\textsuperscript{1} The author is with the Florida Institute of Human and Machine Cognition, 40 S Alcaniz St, Pensacola, FL 32502, United States}}
\thanks{\footnotesize{\textsuperscript{2} The author is with the Department of Mechanical Engineering, Virginia Tech, 635 Prices Fork Rd, Blacksburg, VA 24061, United States}}
\thanks{\scriptsize{Email : \url{ {rgriffin, sbertrand, gwiedebach, jpratt}@ihmc.us}, \url{{leonessa@vt.edu}}}
}}

\def\mytitle{Capture Point Trajectories for Reduced Knee Bend using Step Time Optimization}
\def\mycolumns{twocolumn}
\def\mydocclass{conference}
\def\myabstract{abstract}
\def\mypackages{packages}
\def\bibliocommand{\bibliography{mybib}}

\ifx\mydocclass\undefined
\def\mydocclass{conference}
\fi

\ifx\mycolumns\undefined
\def\mycolumns{onecolumn}
\fi

\ifx\mysize\undefined
\def\mysize{10pt}
\fi

\ifdefined\myinnermargin

\fi
\ifdefined\myoutermargin

\fi
\ifdefined\mytopmargin

\fi
\ifdefined\mybottommargin

\fi

\documentclass[\mysize,\mydocclass,\mycolumns]{ieeeconf}
\IEEEoverridecommandlockouts
\usepackage{graphics}
\usepackage{graphicx}
\usepackage{url}
\usepackage{hyperref}
\usepackage{apalike}
\usepackage{tabulary}
\usepackage{booktabs}

\ifx\bibliocommand\undefined
\else

\usepackage[
    backend=biber,
    natbib=true,
    style=ieee,
    url=false,
    doi=false,
    isbn=false,
    safeinputenc=true
]{biblatex}
\AtEveryBibitem{\clearfield{issn}}
\renewcommand{\bibliography}[1]{\addbibresource{#1.bib}}
\bibliocommand

\fi

\ifdefined\mypackages
\input{\mypackages}
\fi

\renewcommand\[{\begin{equation}}
\renewcommand\]{\end{equation}}

\title{\mytitle}

\ifx\myassociation\undefined
\author{\myauthor}
\else
\author{\myauthor \thanks{\myassociation}}
\fi

\begin{document}

\maketitle
\ifx\myabstract\undefined
\else
\begin{abstract}
\input{\myabstract}
\end{abstract}
\fi

\ifx\mykeywords\undefined
\else
\begin{IEEEkeywords}
\mykeywords
\end{IEEEkeywords}
\fi

\section{Introduction}
\label{introduction}

Humanoid robots have been demonstrated capable of robustly walking across flat surfaces, recovering from pushes and uncertainties, and walking across varying terrain.
They are also demonstrating increasingly dynamic behavior when walking, a major step forward from the slow, quasi-static walking that was previously standard.
However, while there are some exceptions, most all of the gaits employed feature highly bent knees, walking with an almost ``squatted'' motion.
This is highly unnatural, and, while it offers some control benefits, results in significant increases in power consumptions of the knee~\citep{Kajita_2003} over that seen in humans~\citep{Winter_2009}.
It additionally decreases ground clearance of the robot and creates clearance issues when walking over varied terrain.

Before tackling the host of control challenges introduced by walking with straighter legs, the underlying dynamic plans used by the robot must be conducive to walking with less-bent knees.
While whole-body planning schemes that consider the full rigid-body dynamics of the robot have made significant improvements in speed, they typically require too long of solve times for online implementation.
Instead, a typical approach for dealing with the highly nonlinear and high dimensional problem of planning and controls for humanoids is to restrict consideration to the center of mass (CoM).
These low-order dynamic representations have been often preferred for online implementation due to their speed and efficiency.
The zero-moment point (ZMP) is an often utilized representation of the CoM dynamics, and is the point on the ground plane where the moment induced by the inertial and gravitational force is perpendicular to the surface~\citep{Sardain_2004},
It has been used to generate stable walking motions that avoid tipping about the foot edges of the robot~\citep{Kajita_2003, Harada_2006, Wieber_2006}.
Alternatively, the instantaneous capture point (ICP)~\citep{Pratt_2006} and divergent component of motion (DCM)~\citep{Takenaka_2009} were introduced as stable state transformations of the CoM, and have been shown as highly effective strategies for momentum-based planning and control of humanoid robots~\citep{Englsberger_2014, Hopkins_2014}.

Several works have attempted to address planning and control of appropriate height trajectories for straighter legs.
In~\citep{Li_2010}, the authors presented an approach for generating straightened-knee trajectories by combining model predictive control (MPC) of the ZMP to generate horizontal CoM trajectories with a spring-damper approach to generate vertical CoM trajectories.
Alternatively,~\citep{Brasseur_2015} used linear differential inclusion to incorporate these height trajectories directly into the MPC, resulting in relatively natural, cyclical motions of the CoM.
In~\citep{Heerden_2015}, the authors also included CoM height in their ZMP-based quadratic program, but required sequential quadratic programming to solve the resulting problem's quadratic constraints.
The planning and control of height trajectories as they affect the DCM was addressed in~\citep{Hopkins_2014}, which relaxed the assumption of a constant pendulum height in an attempt to increase control of the CoM height over variable terrain.
This last is an example of the standard planning approach, as noted by~\citep{Brasseur_2015}, with CoM height planning typically done independently of horizontal planning, leaving planners ill-equipped to deal with kinematic limitations.
This issue comes to the forefront when attempting to walk with straighter legs.

\begin{figure}[!t]
\centering
\includegraphics[width=3in]{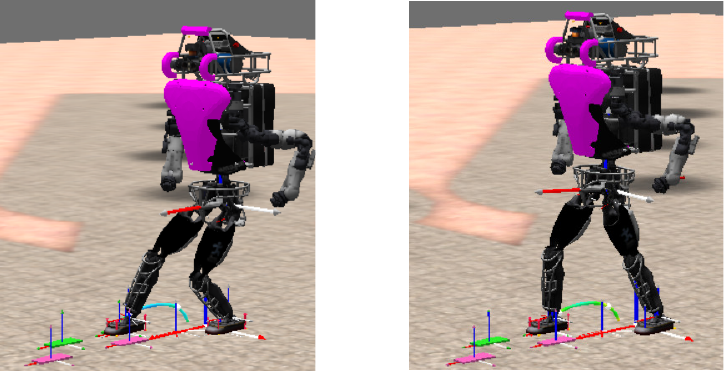}
\vspace{-4mm}
\caption{
Without considering the effects of step timing, ICP plans can require significant knee-bend to execute (left). 
Modifying the timing, however, can result in dynamic, natural, and efficient dynamic plans (right).
}
\vspace{-6mm}
\label{fig:effects_of_timing}
\end{figure}

While some previous works are capable of generating kinematically feasible height trajectories~\citep{Li_2010, Brasseur_2015}, these approaches, along with most others, typically rely on arbitrarily chosen single and double support phase durations.
While kinematic feasibility of the CoM height plan is guaranteed, this has the fundamental limitation of ignoring the effects of step timing on the resulting horizontal CoM plans.
This is, in turn, ignoring the kinematic constraints placed on the system by the timing and the dynamic plan, potentially requiring significant knee-bend during execution.
As the walking speed of most humanoid robot platforms is still relatively slow, this becomes common, with the robot needing to crouch for the next step to be kinematically feasible, as shown on the left of \autoref{fig:effects_of_timing}.
As the step length increases, the required knee-bend of this slow step correspondingly increases, as well.
To address this, we present in this work an approach for optimizing the swing and transfer durations to produce ICP trajectories that do not require knee-bend beyond an amount specifiable by the operator.
We utilize a novel quadratic program that uses the gradient of the ICP plan w.r.t. to time to compute the required timing adjustments that satisfy the desired maximum and minimum knee-bend kinematic constraints.
We believe that proper timing selection is critical to the continued progress of humanoid locomotion towards efficient, human-like gaits.

\section{Dynamic Planning Background}
\label{dynamicplanningbackground}

An increasingly common approach for dynamic planning in humanoid robots is to utilize a stable transformation of the CoM position and velocity in the form of either the ICP, DCM, or extrapolated center of mass (XCoM)~\citep{Pratt_2006, Takenaka_2009, Pratt_2012, Morisawa_2012, Ramos_2014, Hopkins_2014, Hof_2005, Englsberger_2014}.
In this work, we will refer only to the ICP, noting its equivalency to the DCM and XCoM in the $x$-$y$ plane.
The ICP is, as mentioned, a stable transformation of the CoM state, and is defined as
\[
\V{\xi} = \V{x} + \tfrac{ 1 }{ \omega } \dot{\V{x}},
\label{eqn:icp_definition}
\]
where $\V{\xi}$ is the ICP position, $\V{x}$ and $\dot{\V{x}}$ are the CoM position and velocity, and $\omega = \sqrt{g / \Delta z_{\text{com}}}$ is the natural frequency of the inverted pendulum. By reordering this, we can see that the CoM has stable first order dynamics with respect to the ICP, meaning that it will converge to the ICP over time.
Through differentiation, the ICP dynamics are defined as
\[
\dot{\V{\xi}} = \omega \left( \V{\xi} - \V{r}_{\text{cmp}} \right),
\label{eqn:icp_dynamics}
\]
where we see that the Centroidal Moment Pivot (CMP) point~\citep{Popovic_2005}, $\V{r}_{\text{cmp}}$, controls the ICP dynamics.

\subsection{Dynamic Planning of ICP Trajectories}
\label{dynamicplanningoficptrajectories}

There are a variety of methods that have been used to generate stable ICP trajectories, such as~\citep{Hopkins_2014}, where discrete trajectories were generated numerically from specified ZMP trajectories.
In this work, we use the algorithm proposed in~\citep{Englsberger_2014}, summarized in the following paragraphs.

\begin{figure}[!t]
\vspace{2mm}
\centering
\includegraphics[width=3in]{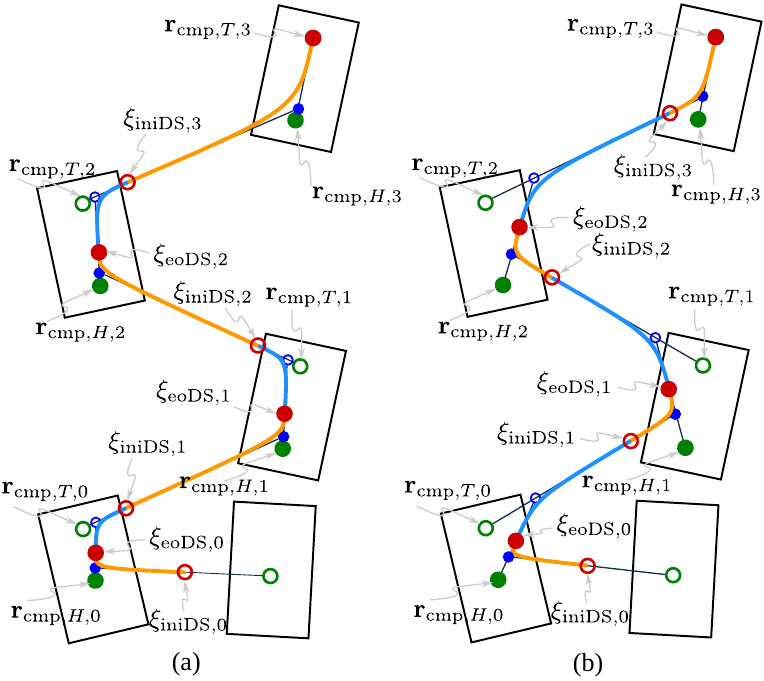}
\vspace{-4mm}
\caption{Heel-to-Toe ICP trajectory with smooth CMP trajectories for slower (left) and faster (right) steps~\citep{Englsberger_2014}.}
\vspace{-6mm}
\label{fig:recursive_icp_plan}
\end{figure}

From the definition of the ICP dynamics in \autoref{eqn:icp_dynamics}, the differential equation has an analytic solution
\[
\V{\xi}(t) = e^{\omega t} \left( \V{\xi}_0 - \V{r}_{\text{cmp}} \right) + \V{r}_{\text{cmp}},
\label{eqn:icp_dynamics_solution}
\]
assuming $\V{r}_{\text{cmp}}$ is held constant.
Using this, we can calculate a desired ICP trajectory for walking, given a set of desired footsteps and desired CMP locations in those footsteps.
To more accurately represent human-like walking, we use two CMPs per foot, one in the heel ($\V{r}_{\text{cmp},H}$) and one in the toe ($\V{r}_{\text{cmp},T}$), as shown in \autoref{fig:recursive_icp_plan} by the green circles.
This results in the reference CMP trajectory moving from the heel to the toe in the foot while stepping.

To determine the amount of time spent on the CMP, we can break the ICP plan into four segments, $T_{\text{iniDS}}, T_{\text{endDS}}, T_{\text{iniSS}},$ and $T_{\text{endSS}}$.
These correspond roughly to the time in transfer shifting the weight to the upcoming support foot, the time in transfer shifting the weight forward in the foot, the time in swing shifting the weight forward in the foot, and the time spent in swing with the weight in the front of the foot, respectively.
The desired ICP trajectory can be calculated by recursing backward from the final objective location.
This can be done by using the solution to the ICP dynamics in \autoref{eqn:icp_dynamics_solution}, and assuming a static CMP location, allowing the ICP locations to be computed at the beginning and end of swing and transfer.
We can smooth the ICP trajectories using third order polynomial interpolation between each of these points, which guarantees smoothness of the CMP trajectory~\citep{Englsberger_2014}, resulting in the light blue and orange colored lines in \autoref{fig:recursive_icp_plan}.

Note that the effect of taking slower steps is highlighted in \autoref{fig:recursive_icp_plan}.
This results in much more of the ICP motion occurring during transfer.
This, in turn, has great effects on the location of the CoM when walking.

\subsection{Computation of CoM Trajectory Solutions}
\label{computationofcomtrajectorysolutions}

While the utilization of the ICP for dynamic planning has the major advantage of not requiring specific calculation of CoM trajectories, this is required for determination of kinematic feasibility.
In particular, we want to know the location of the CoM at touchdown, as this is the farthest point from both the leading and trailing feet.
By using the ICP algorithm in~\citep{Englsberger_2014}, we can obtain an analytic solution for the CoM trajectory.
If the ICP trajectory is defined using a constant CMP, as in \autoref{eqn:icp_dynamics_solution}, the CoM trajectory can be defined by integrating the CoM dynamics
\[
\dot{\V{x}}(t) = \omega \left( e^{\omega t} \left( \V{\xi}_0 - \V{r}_{\text{cmp}} \right) + \V{r}_{\text{cmp}} - \V{x}(t) \right),
\]
which has the solution
\[
\begin{aligned}
\V{x}(t) = 
\tfrac{1}{2} e^{\omega t} \left( \V{\xi}_0 - \V{r}_{\text{cmp}} \right) - 
\tfrac{1}{2} e^{-\omega t} \left( \V{\xi}_0 + \V{r}_{\text{cmp}} - 2 \V{x}_0 \right) 
+ \V{r}_{\text{cmp}}.
\end{aligned}
\label{eqn:com_dynamic_trajectory}
\]
Alternatively, if the ICP trajectory is defined using a cubic polynomial, the CoM dynamics become
\[
\dot{\V{x}}(t) = \omega \left( \V{c}_0 + \V{c}_1 t + \V{c}_2 t^2 + \V{c}_3 t^3 - \V{x}(t) \right),
\]
which has the solution
\[
\begin{aligned}
\V{x}(t) = & \V{c}_3 t^3 + \left( \V{c}_2 - \tfrac{3}{\omega} \V{c}_3 \right) t^2 + \left( \V{c}_1 - \tfrac{2}{\omega} \V{c}_2 + \tfrac{6}{\omega^2} \V{c}_3 \right) t \\
& + \left( \V{c}_0 - \tfrac{1}{\omega} \V{c}_1 + \tfrac{2}{\omega^2} \V{c}_2 - \tfrac{6}{\omega^3} \V{c}_3 \right) \\
& + e^{-\omega t} \left( \V{x}_0 - \V{c}_0 + \tfrac{1}{\omega} \V{c}_1 - \tfrac{2}{\omega^2} \V{c}_2 + \tfrac{6}{\omega^3} \V{c}_3 \right).
\label{eqn:com_polynomial_trajectory}
\end{aligned}
\]
Using these equations, we can then calculate where the CoM will be located at touchdown.

\section{Center of Mass Adjustment Calculation}
\label{centerofmassadjustmentcalculation}

\begin{figure}[!t]
\vspace{2mm}
\centering
\includegraphics[width=2.8in]{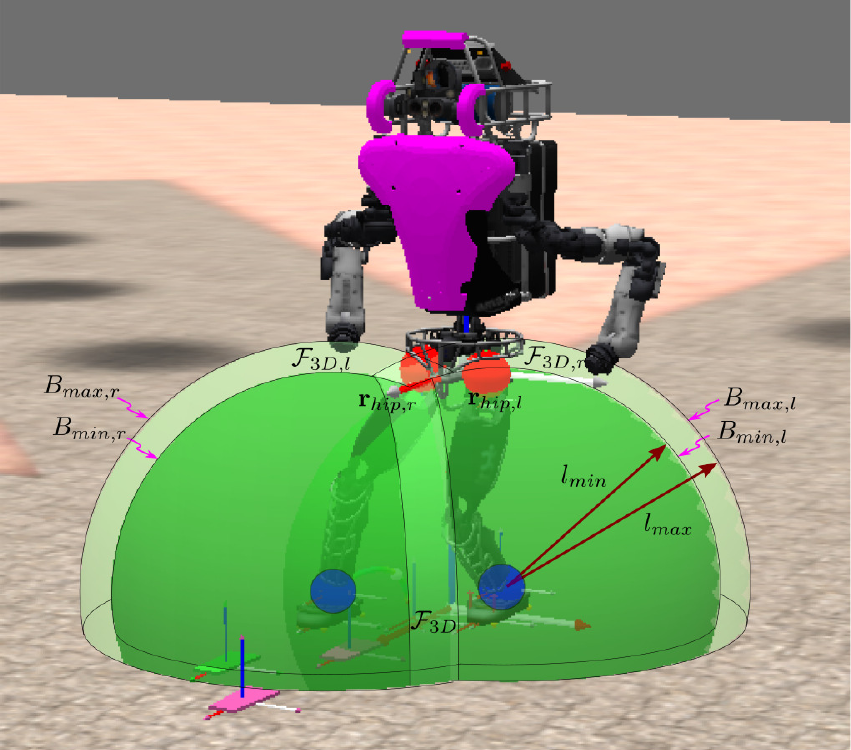}
\caption{
Illustration of the kinematic constraints we are using to calculate the desired CoM adjustments.
The light green spheres represent the area in which the hips (the red dots) must be located to be reachable with the legs at maximum extension.
The dark green spheres represent the area outside of which the hips must be located to guarantee bending less than some specifiable amount.
}
\vspace{-6mm}
\label{fig:reachability_illustration}
\end{figure}

Before we can calculate the desired timing adjustments in the ICP plan, we must determine \emph{how} we want to adjust the plan itself.
To do this, we can determine the requirements placed on the CoM location that satisfy our kinematic constraints placed on the system; namely that of maximum and minimum leg extension.
We can use these constraints to define the ``feasibility region'' for the CoM as the area where the CoM can be located so that both legs do not bend more or less than desired. 

To compute the three dimensional feasibility region, we can observe that placing limits on the amount of knee bend at both full retraction and full extension acts to constrain the location of the robot's hips.
We can calculate these areas exactly by defining a series of spheres whose radius is a function of the retracted (when it is most bent) and extended (when it is least bent) leg length.
By defining a maximum and minimum amount of leg bend, $\theta_{\text{max}}$ and $\theta_{\text{min}}$, respectively, we can calculate the corresponding bent and extended leg length, $l_{min}$ and $l_{max}$.
We can then use this to define a sphere centered at the ankle, $B_{\text{min}}$ that the hip joint must be located outside of, shown by the dark green spheres in \autoref{fig:reachability_illustration}.
In doing so, we then know that as long as, at touchdown, both hip joints are located outside of these spheres, the ICP plan will not require either knee to bend past $\theta_{\text{max}}$.
Additionally, a second set of spheres can be defined, $B_{\text{max}}$, shown by the lighter green spheres in \autoref{fig:reachability_illustration}, to constrain the hip locations to be achievable at full leg extension.

Using these concentric set of spheres, we can compute the 3D feasibility region, $\mathcal{F}_{3D} \subset \mathbb{R}^3$, to constrain the CoM location by saying the hip locations must be within $B_{\text{max}}$ and outside of $B_{\text{min}}$, or
\[
\begin{aligned}
\V{r}_{\text{hip}} \in \mathcal{F}_{3D} \equiv \left\{ B_{\text{max}} - \ B_{\text{min}} \right\} \subset \mathbb{R}^3,
\end{aligned}
\label{eqn:reachability_sphere}
\]
which is illustrated in \autoref{fig:reachability_illustration}.
Here, the ankle locations are the blue balls, while the hip locations of the robot are the red balls.
As can be seen, if both hips are located inside the sphere of the maximum leg length and outside the sphere of the minimum leg length at touchdown, the plan is both kinematically achievable and will not require bending either knee past $\theta_{\text{max}}$.

\begin{figure}[!t]
\vspace{2mm}
\centering
\includegraphics[width=3.4in]{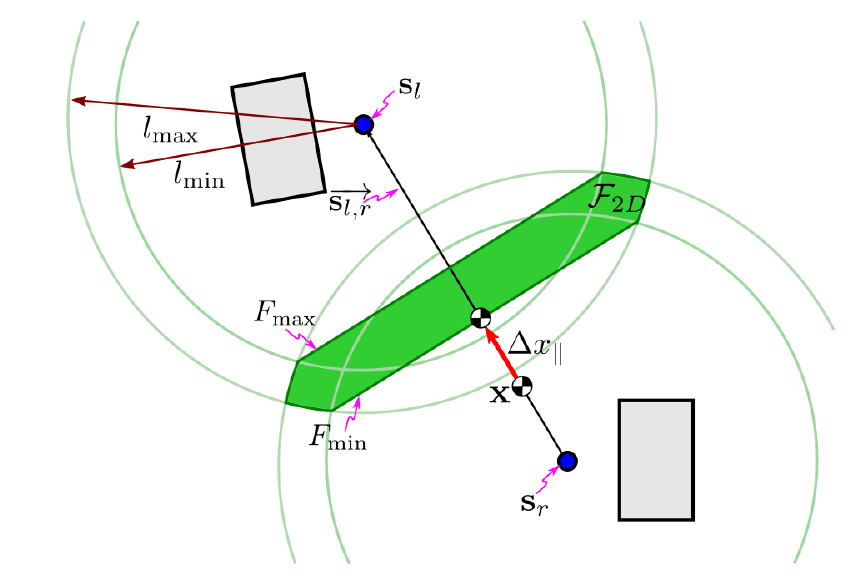}
\vspace{-8mm}
\caption{
Illustration of the approach for computing the desired CoM adjustment. 
The region where the CoM is both kinematically feasible and satisfies the max knee bend is represented by the shaded green area.
The desired CoM adjustment is the distance from CoM to the feasibility region.
}
\vspace{-6mm}
\label{fig:reachability_adjustment}
\end{figure}

From the 3D feasible space, we can compute the 2D feasibility region, $\mathcal{F}_{2D} \subset \mathbb{R}^2$, in the $x$-$y$ plane where the CoM must be located, shown as the shaded green region in \autoref{fig:reachability_adjustment}.
To do so, we assume that the distance from the CoM to each hip can be treated as a constant, a fairly common~\citep{Li_2010} and accurate assumption due to the mass distribution on most humanoid robots.
This allows us to offset the sphere origins from the ankle locations, defining the blue circles in \autoref{fig:reachability_adjustment} as $\V{s}_{l}$ and $\V{s}_r$, enabling us to directly consider the CoM rather than the hip locations as in \autoref{fig:reachability_illustration}.
$\mathcal{F}_{2D}$ then becomes much simpler to calculate, being bounded on the sides by straight lines perpendicular to the vector from one ankle to the other when walking with constant ankle height,
\[
\begin{aligned}
F_{\text{min}} \perp \overrightarrow{\V{s}_{l,r}}, & \ \ & 
F_{\text{max}} \perp \overrightarrow{\V{s}_{l,r}}.
\end{aligned}
\label{eqn:line_bounds}
\]
This follows directly from projecting $\mathcal{F}_{3D}$ in \autoref{eqn:reachability_sphere} onto the $x$-$y$ plane.
This approach extends to non-flat ground by using arcs instead of lines.

The objective then becomes to compute the distance required to project the CoM into the feasibility region,
\[
\begin{aligned}
\Delta x_\parallel = \text{d} \left(\V{x}, \mathcal{F}_{2D} \right).
\end{aligned}
\]
We can simplify this adjustment as being along the vector $\overrightarrow{\V{s}_{l,r}}$, the red line in \autoref{fig:reachability_adjustment}.
From the definition of the bounds of the set in \autoref{eqn:line_bounds}, this is equivalent to
\[
\begin{aligned}
\Delta x_\parallel = 
\begin{cases}
\text{d}(\V{x}, F_{\text{max}}), & x > f_{\text{max}}, \\
0, & f_{\text{min}} \le \V{x} \le f_{\text{max}}, \\
\text{d}(\V{x}, F_{\text{min}}), & x < f_{\text{min}},
\end{cases}
\end{aligned}
\]
where $f_{\text{max}}$ and $f_{\text{min}}$ are the points where $\overrightarrow{\V{s}_{l,r}}$ intersections $F_{\text{max}}$ and $F_{\text{min}}$, respectively.
This provides the desired CoM adjustment of ICP plan to achieve the desired kinematic motions.

\section{Timing Optimization Algorithm}
\label{timingoptimizationalgorithm}

Now that we can compute the desired CoM adjustment, we must develop a tool for modifying the ICP plan to achieve this adjustment.
The CoM trajectory is a function of the CMP position and step timing.
As the CMP position is defined according to the step plan, it can be assumed fixed, requiring the durations used in the ICP plan to be modified to achieve the necessary CoM adjustment.
The objective CoM adjustment can be found by computing the CoM location at touchdown, $\V{x}_f$, at the beginning of each step through integrating the dynamic plan.
Using the planning algorithm in \autoref{dynamicplanningbackground}, this can be done with the analytic solutions in \autoref{eqn:com_dynamic_trajectory} and \autoref{eqn:com_polynomial_trajectory}, although it can be found using numerical integration for any ICP plan, as well.
Then, using $\V{x}_f$, we can compute the objective parallel adjustment $\Delta x_\parallel$ using the approach outlined in \autoref{centerofmassadjustmentcalculation}.
The ICP plan however, is a function of several timing elements, meaning the timing adjustments cannot be explicitly solved for from only the desired CoM adjustment, as the problem is under-constrained.
Instead, we can cast the problem in an optimization framework, defining additional cost objectives to calculate an optimal set of timing adjustments.
We will do so in the following section by defining a quadratic program (QP) with an iterative feedback loop.

From the definition of the ICP plan in \autoref{dynamicplanningbackground}, we can define the step timing variables to be used in the adjustment calculation as
\[
\begin{aligned}
\V{T} = 
\left[
\begin{array}{cccc}
T_{\text{iniDS},0} & T_{\text{endDS},0}    &    T_{\text{iniSS},0}  & \dots
\end{array}
\right.
\\
\left.
\begin{array}{ccc}
T_{\text{endSS},0}     &     T_{\text{iniDS},1}    &    T_{\text{endDS},1}
\end{array}
\right]^T,
\end{aligned}
\label{eqn:cost_function_variables}
\]
where the subscript number indicates the associate step, with $0$ being current.
While these variables are specific to the ICP planning approach presented in~\citep{Englsberger_2014}, variables for other ICP planning approaches can be equivalently defined.
Due to the definition of the ICP dynamics, however, the ICP and resulting CoM plans are highly nonlinear with respect to time, where the CoM location at touchdown can be defined as some nonlinear function,
\[
\begin{aligned}
\V{x}_f = \V{f}(\V{T}).
\end{aligned}
\]
A method of determining the optimal durations, $\V{T}^*$ is required.
While nonlinear optimization is possible, we prefer to use a QP due to its reliability and efficiency.
To do so, we can approximate the gradients of $\V{f}(\cdot)$ with respect to $\V{T}$,\\
 \[
 \begin{aligned}
\nabla_{\V{T}} \V{f} = 
\left[
\begin{array}{cccc}
 \nabla \V{f}_{T_{\text{iniDS},0}} &
 \nabla \V{f}_{T_{\text{endDS},0}} &
 \nabla \V{f}_{T_{\text{iniSS},0}} & 
 \dots
 \end{array}
 \right. \\
 \left.
 \begin{array}{ccc}
 \nabla \V{f}_{T_{\text{endSS},0}} &
 \nabla \V{f}_{T_{\text{iniDS},1}} &
 \nabla \V{f}_{T_{\text{endDS},1}}
 \end{array}
 \right],
 \end{aligned}
 \label{eqn:gradient_function}
 \]
 by slightly perturbing the function $\V{f}(\cdot)$ such that
 \[
 \begin{aligned}
 \delta \V{x}_f = \nabla_{\V{T}} \V{f} \delta \V{T} + \text{H.O.T.}.
 \end{aligned}
 \label{eqn:gradient_dynamics}
 \]
The gradient function in \autoref{eqn:gradient_function} can be broken into components parallel and perpendicular to the desired CoM adjustment, $\nabla f_{\parallel} \left( \V{T} \right)$ and $\nabla f_{\perp} \left( \V{T} \right)$, respectively.
This then allows us to define a QP using $\nabla_{\V{T}} \V{f}(\V{T})$.

\begin{figure*}[!t]
\centering
\vspace{2mm}
\includegraphics[width=7in]{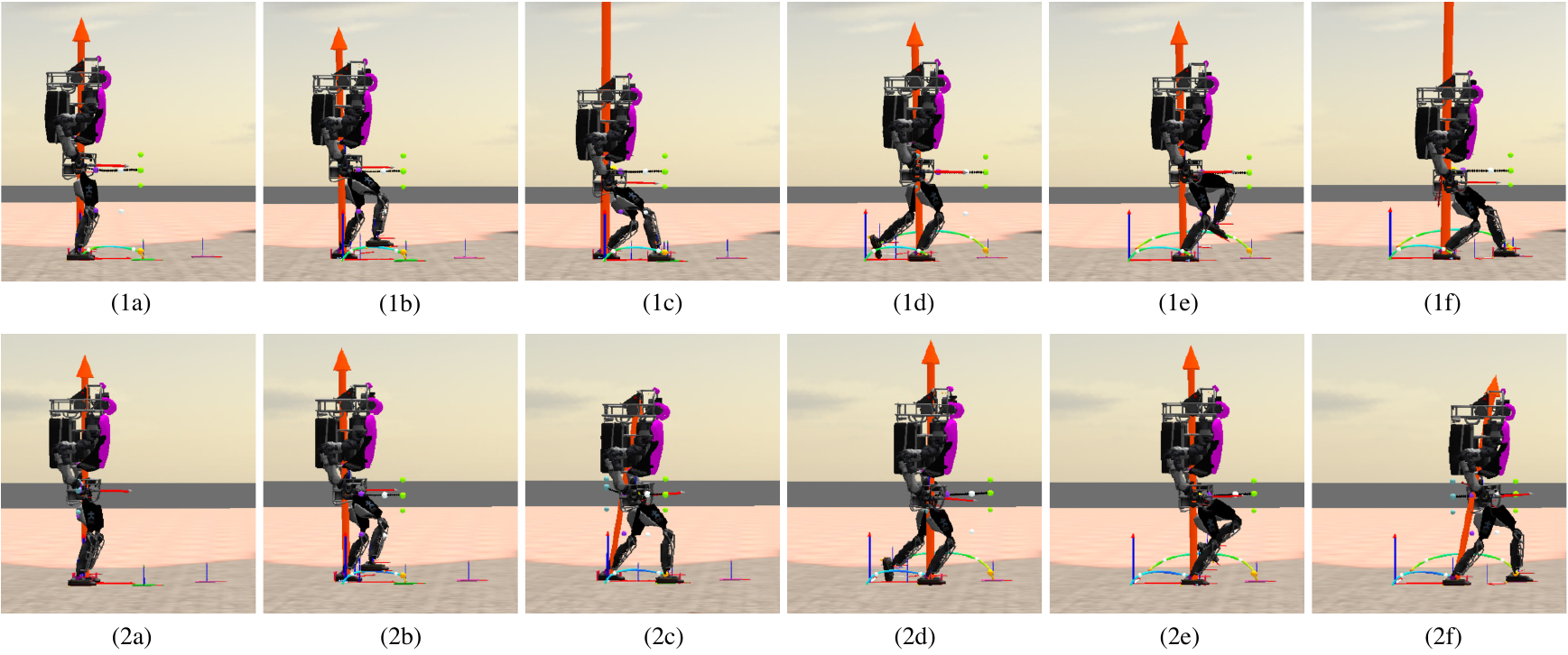}
\vspace{-8mm}
\caption{
Comparison between walking without the time optimization module active (top) and with the time optimization module active (bottom).
The desired steps are 0.6m long, using a 2.5s swing and 2.5s transfer.
In this case, the maximum desired knee bend is 0.4 rads, and the resulting total step duration is reduced.
}
\vspace{-8mm}
\label{fig:walking_time_lapse}
\end{figure*}

We can describe the desired CoM adjustment as a quadratic objective function of the step durations,
\[
\begin{aligned}
 J_\parallel = \left\| \Delta x_{\parallel} - \nabla f_{\parallel} \Delta \V{T} \right\|^2_{Q_{\parallel}}.
\end{aligned}
\]
We prefer to define this as an objective instead of an equality constraint, as this formulation prevents the problem from being over-constrained if limits are placed on $\Delta \V{T}$.
We can then define additional quadratic objective function, such as the minimization of perpendicular CoM adjustments,
\[
\begin{aligned}
J_\perp = \left\| \nabla f_{\perp} \Delta \V{T} \right\|^2_{Q_{\perp}},
\end{aligned}
\]
and the minimization of the total timing adjustment,
\[
\begin{aligned}
J_T = \left\| \Delta \V{T} \right\|^2_{\V{R}_T}.
\end{aligned}
\]
Lastly, we can define a cost function to minimize the timing symmetry between the duration of the beginning of each walking phase, $T_{\text{ini},i}$, and the end of each walking phase, $T_{\text{end},i}$,\\
\[
\begin{aligned}
J_\alpha = \left\| \Delta \V{T}_{\text{ini}} - \Delta \V{T}_{\text{end}}, \right\|_{R_\alpha}^2,
\end{aligned}
\]
which yields ICP plans with more desirable CMP trajectories when using the planning approach from~\citep{Englsberger_2014}.
The total cost function is then defined as
\[
\begin{aligned}
J = J_\parallel + J_\perp + J_T + J_\alpha, 
\end{aligned}
\]
with a corresponding optimal solution
\[
\begin{aligned}
\Delta \V{T}^* = \text{argmin} \ J(\Delta x_\parallel, \Delta \V{T}).
\end{aligned}
\]

\begin{figure}[!t]
\centering
\vspace{2mm}
\includegraphics[width=3.4in]{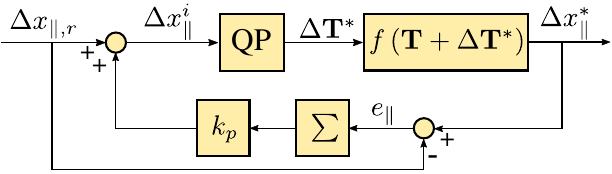}
\vspace{-5mm}
\caption{
Block diagram of the feedback loop for calculating the timing adjustment.
The loop is performed until the desired adjustment is achieved, with a feedback based on the achieved adjustment in each iteration.
}
\vspace{-6mm}
\label{fig:block_diagram}
\end{figure}

However, as this algorithm ignores the higher order terms in \autoref{eqn:gradient_dynamics}, the resulting CoM adjustment from $\Delta \V{T}^*$ are not likely to be equivalent to the desired adjustment, $x_\parallel \ne f_\parallel ( \V{T} + \Delta \V{T}^*)$.
This means that the QP is not likely to find an adjustment satisfying the kinematic constraints on the first iteration.
To compensate for this, we can, similar to Sequential Quadratic Programming, define an iterative feedback-based loop
\[
\begin{aligned}
\Delta x_\parallel^{i+1} = \Delta x_\parallel^i + k_p e_\parallel,
\end{aligned}
\]
where $e_\parallel = \Delta x_{\parallel,r} - \Delta x_\parallel^*$ is the error between the achieved adjustment using the QP solution and the original desired CoM adjustment, $\Delta x_\parallel^i$ is the desired adjustment used in the current solver iteration, and $\Delta x_\parallel^{i+1}$ is the desired adjustment for the next iteration.
This feedback loop is repeated until the error $e_\parallel$ is less than a fixed amount, $| e_\parallel | < \epsilon$, at which the loop is terminated.
At the end of each iteration, the objective adjustment for next iteration, $\Delta x_\parallel^{i+1}$, is computed.
The loop can also be terminated once a specified maximum number of iterations is exceeded.

\section{Results}
\label{results}

The algorithm was implemented in the IHMC Open Source Software system, using the Simulation Construction Set.
To solve the QP, a custom implementation of the Goldfarb-Idnani solver was used.
Balance is controlled when simulating the Atlas humanoid using the momentum-based whole-body controller outlined in~\citep{Koolen_2013}.

\begin{figure}[!t]
\vspace{2mm}
\centering
\includegraphics[width=3.3in]{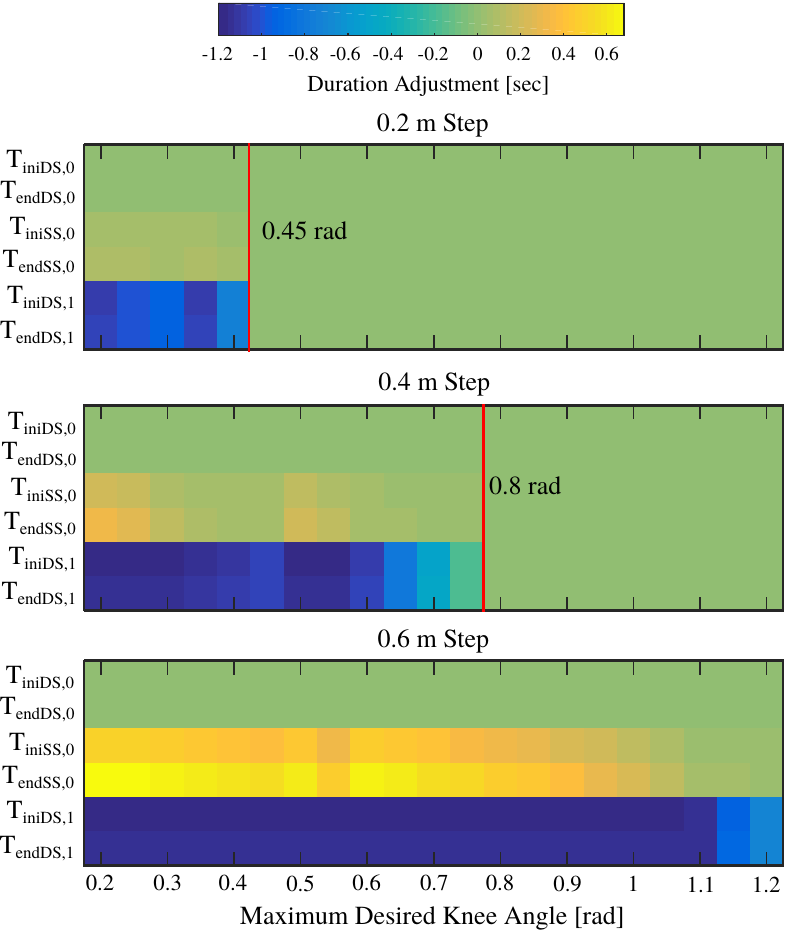}
\vspace{-3mm}
\caption{
Timing adjustments while taking 0.2m, 0.4m, and 0.6m steps and 5s pre-adjusted step durations.
}
\vspace{-6mm}
\label{fig:timing_adjustment_results}
\end{figure}

A simulation of walking using \unit{5.0}{s}, \unit{0.6}{m} steps with and without using this algorithm is shown in \autoref{fig:walking_time_lapse}.
As can be seen in the top row, specifically panels 1c and 1f, using the fixed provided timing without adjustment requires significant bending of the knee for the step to be reached.
The bottom row illustrates walking using the presented algorithm.
The step durations are modified to require only \unit{0.4}{rad} of knee-bend, producing a much more natural, dynamic gait.
The reduction in knee bend can be seen when comparing with panels 2c and 2f.
The necessary timing adjustments are shown in \autoref{fig:timing_adjustment_results}, and are discussed in detail later.

The resulting timing adjustments calculated using the algorithm for three different step lengths to achieve different degrees of knee bend are shown in \autoref{fig:timing_adjustment_results}.
Here, we compare the timing adjustment results for relatively slow steps, a \unit{2.5}{s} swing duration and \unit{2.5}{s} transfer duration when allowing a differing degree of knee-bend.
For short (\unit{0.2}{m}) steps, the motion results in approximately \unit{0.45}{rad} of bend at the knee using the original, slow \unit{5.0}{s} step duration.
If we specify that we would like the legs to be straighter, i.e. less knee-bend, the transfer duration of the upcoming step is decreased until the knees are only bent by the specified amount, while the other durations are unchanged.
The optimization prefers to use the upcoming transfer duration, as it is the most effective means to adjust the CoM position, as illustrated in \autoref{fig:recursive_icp_plan}.

For medium length (\unit{0.4}{m}) steps, as shown in the middle plot of \autoref{fig:timing_adjustment_results}, the step requires significantly more knee-bend at the original \unit{5}{s} duration than the short step length, needing approximately \unit{0.8}{rad} of bend.
The optimization again primarily utilizes the upcoming transfer duration.
As we placed constraints on the minimum durations for the different transfer segments, however, the current swing duration is also somewhat adjusted in the most extreme cases, as this offers slight modifications of the CoM positions.

For longer (\unit{0.6}{m}) steps, as shown in the bottom plot of \autoref{fig:timing_adjustment_results}, the user specified step duration of \unit{5}{s} requires a large amount of knee-bend, greater than \unit{1.2}{rad}.
By trying to minimize knee bend, the upcoming transfer duration is made as small as possible.
The current swing duration is also increased as the maximum knee bend is decreased in an attempt to achieve the desired amount of CoM adjustment.

It is worth noting that, for all step lengths, the current transfer duration has virtually no effect on the final CoM position when using the ICP planner outlined in~\citep{Englsberger_2014}.
Additionally, in practical implementation, allowing the current swing duration to increase provides relatively little benefit, with almost all effective adjustments coming from the upcoming transfer duration.

\section{Conclusion}
\label{conclusion}

Using the presented approach, we were able to modify existing ICP plans to produce walking gaits that require minimal knee bend.
Traditional walking approaches utilize highly bent knees, which has been somewhat addressed in other works by attempting to straighten the legs to their maximum feasible length while walking.
This, however, overlooks the kinematic constraints that fixed step timing places on the CoM height and corresponding knee angles.
This work represents, to our knowledge, the first study of the effects of the step duration in the resulting dynamic plans and their corresponding CoM heights.

We presented an approach that utilizes a quadratic program optimization scheme to compute the necessary step timing adjustments to produce plans that require a specified amount of knee bend, while remaining kinematically feasible.
This algorithm is then illustrated to successfully adjust the ICP plan timing to reduce the knee bend to the desired angle.
As robots move towards walking with straighter legs and more natural, efficient gaits, the effects of step timing will become increasingly important, with this work representing a critical first step towards achieving these motions.

In future work, we plan to utilize this algorithm along with advanced CoM height control techniques to enable humanoid robots to efficiently walk with straight legs using momentum-based whole-body control frameworks.
We also plan to extend this work to account for kinematic effects of toe-off and pelvis orientation changes on reachability.

\input{footer.tex}

\end{document}

%% file: packages.tex
\usepackage{array}
\usepackage{amsmath}
\usepackage{amsfonts}
\interdisplaylinepenalty=2500
\usepackage{algorithm}
\usepackage{algpseudocode}
\usepackage{units}

\usepackage{dirtytalk}
\usepackage{xcolor}
\usepackage[english]{babel}

\input{defs}

\linespread{1.0}

%% file: defs.tex
\newcommand{\V}[1] {\boldsymbol{\mathbf{#1}}}


\def\IEEEeqnarraybox{\@IEEEeqnarraystarformfalse\ifmmode\@IEEEeqnarrayboxHBOXSWfalse\else\@IEEEeqnarrayboxHBOXSWTRUE\fi%
\@IEEEeqnarraybox}
\def\endIEEEeqnarraybox{\end@IEEEeqnarraybox}

\newcommand{\arraybegin}[1]{\begin{IEEEeqnarraybox*}[][c]{#1}}
\newcommand{\arrayend}{\end{IEEEeqnarraybox*}}

%% file: abstract.tex
Traditional force-controlled bipedal walking utilizes highly bent knees, resulting in high torques as well as inefficient, and unnatural motions.
Even with advanced planning of center of mass height trajectories, significant amounts of knee-bend can be required due to arbitrarily chosen step timing.
In this work, we present a method that examines the effects of adjusting the step timing to produce plans that only require a specified amount of knee bend to execute.
We define a quadratic program that optimizes the step timings and is executed using a simple iterative feedback approach to account for higher order terms.
We then illustrate the effectiveness of this algorithm by comparing the walking gait of the simulated Atlas humanoid with and without the algorithm, showing that the algorithm significantly reduces the required knee bend for execution.
We aim to later use this approach to achieve natural, efficient walking motions on humanoid robot platforms.

%% file: footer.tex

\ifx\myappendix\undefined
\else

\section*{APPENDIX}
\input{\myappendix}

\fi

\ifx\myacknowledgments\undefined
\else

\section*{ACKNOWLEDGMENT}
\input{\myacknowledgments}

\fi

\ifx\bibliocommand\undefined
\else
\printbibliography
\fi